\documentclass[oneside,11pt]{article}

\usepackage{color}
\usepackage{graphicx}
\usepackage{amsmath}
\usepackage{algorithm}
\usepackage[noend]{algpseudocode}
\usepackage{float}
\usepackage[utf8]{inputenc}

\usepackage[a4paper, total={6in, 9in}]{geometry}
\usepackage[english]{babel}
\usepackage[backend=biber]{biblatex}

\title{Optimizing PID parameters with machine learning}
\author{Adam Nyberg}
\date{ }

\begin{document}

\maketitle

\begin{abstract}
This paper examines the Evolutionary programming (EP) method for optimizing PID parameters. PID is the most common type of regulator within control theory, partly because it's relatively simple and yields stable results for most applications. The p, i and d parameters vary for each application; therefore, choosing the right parameters is crucial for obtaining good results but also somewhat difficult. EP is a derivative-free optimization algorithm which makes it suitable for PID optimization. The experiments in this paper demonstrate the power of EP to solve the problem of optimizing PID parameters without getting stuck in local minimums.
\end{abstract}

\section{Introduction}

This paper will examine one approach to the optimization of Proportional-Integral-Derivative (PID) parameters. PID is the most common type of regulator within control theory. The mathematical expression of a PID regulator is relatively simple and yields stable results for most applications. The p, i and d parameters vary for each application; therefore, choosing the right parameters is crucial for obtaining good results but also somewhat difficult. According to K.J.Åström and T.Hägglund \cite{intro:pid}, there are several methods used in the industry for tuning the p, i and d parameters. Some methods involve developing models while others involve manual tuning by trial and error. 

\vspace{5mm}

This paper examines the Evolutionary programming (EP) method for optimizing PID parameters. EP is an iterative algorithm that runs until some performance criteria are met. The basic steps are as follows: EP generates a population, evaluates every member in the population, selects the best member, creates a new population based on the best member, and then repeats the aforementioned process with an evaluation of the new population. The new population is generated by adding a Gaussian mutation to the parent. The results of this paper are compared with the results generated in R.Johns \cite{rasmus}, which uses Genetic algorithm (GA), and S.Hadenius \cite{simon}, which uses Particle Swarm optimization (PSO). The GA and PSO methods are further explained in \cite{intro:algo}.

\vspace{5mm}

All training and testing of the algorithm was done in a simulated environment called MORSE on a Robot Operating System (ROS). The algorithm and movements of the robot were developed as three ROS nodes. An ROS node is one subsystem; typically, a robotics environment is built with multiple ROS nodes. The actual simulation of the robot was done by MORSE which is a generic 3D simulator for robots. Nodes are only able to communicate with other nodes using streaming topics, Remote Procedure Calls (RPC) services, and the Parameter Server. One route was used for training and a separate route was used for evaluating the step response. The experiments in this paper tuned two PIDs simultaneously. The PIDs tuned were the ones controlling linear velocity and angular velocity on the robot. Only the Husky robot was used.

\vspace{5mm}

EP can perform effectively in this type of domain due to its ability to optimize without getting stuck in local minimums and its efficiency over brute force. The resulting parameters were evaluated by looking at the step response. Important measurements are: rise time, overshoot, and the steady state error.

\section{Method}
\subsection{Set up environment}
This paper describes three different experiments that all ran on the same environment. Everything required to run the experiments is listed in table \ref{table:1}.
 
\begin{table}[ht!]
\centering
\begin{tabular}{| l   p{11cm}|} 
    \hline
     Name & Description \\ [0.5ex] 
     \hline\hline
     Roscore & Roscore is a collection of nodes and programs that serve as prerequisites of a ROS-based system.  \\ 
     \hline
     State machine & The state machine is the ROS node that controls general movement and includes the PID for both linear velocity and angular velocity. This node is built in Modelled Architecture (March).  \\
     \hline
     Route & This node is only responsible for making the robot drive a predefined route. It is triggered by the Determinator, and it sends a callback back to the Determinator when the route is completed. The route used for training can be seen in algorithm \ref{alg:route} with parameters $start = -0.3$ and $end = 0.3$. The route used for testing uses the same algorithm but with parameters $start = 0.1$ and $end = 0.7$. The parameters $start$ and $end$ determine the velocity of the robot. The test route was run with the best PID parameters from the training.\\
     \hline
     Determinator & This node runs the algorithm and logs data to file. This is also the node that starts the route node. A detailed description of the EP algorithm is in section \ref{EPA}.\\
     \hline
\end{tabular}
\caption{Table of systems required for running the experiments.}
\label{table:1}
\end{table}

\begin{algorithm}
\caption{The route used by the robot for train and test.}
\label{alg:route}
\begin{algorithmic}[1]
\Procedure{Route}{start, end}
\State $linearVelocity \gets start$
\State $angularVelocity \gets start$
\State $\text{wait 3 seconds}$
\State $linearVelocity \gets end$
\State $angularVelocity \gets end$
\State $\text{wait 3 seconds}$
\State $\text{Notify finnished to Determinator}$
\EndProcedure
\end{algorithmic}
\end{algorithm}
\vspace{5mm}

\subsection{Definitions}\label{Defs}
The Determinator receives both the actual velocity and desired velocity of the robot 50 times per second. With that data, an error can be estimated by calculating $abs(desired - actual)$. That calculation is executed for every sample, summed, and then divided by the total number of samples. Thus, the total error for one run of a route is the average error of that route. This average error (AE) is used by the algorithm to evolve the parameters. Fitness refers to lowest AE in this paper.

\vspace{5mm}

\begin{figure}[!ht]
   \centering
       \includegraphics[width=0.70\textwidth]{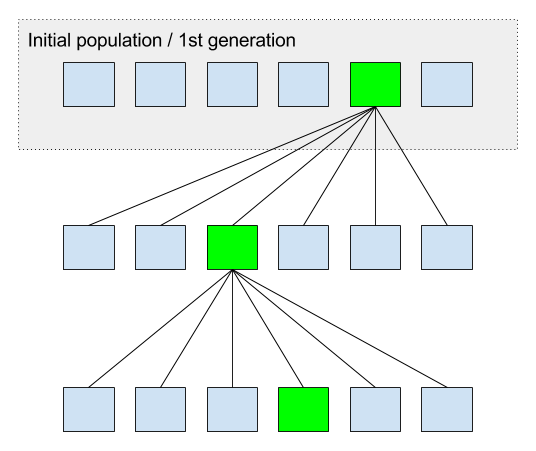}
       \caption{Figure illustrating 3 generations with 6 individuals in each generation.}
   \label{fig:gens}
\end{figure}

A population, or generation, is a collection of individuals in which each individual contains values for $k_{pv}, k_{iv}, k_{dv}, k_{pa}, k_{ia}, k_{da}$. The values denoted $k_{pv}, k_{iv}, k_{dv}$ controls the linear velocity PID and values denoted $k_{pa}, k_{ia}, k_{da}$ controls the angular velocity PID. The velocity and angular parameters were evaluated separately from one another. Figure \ref{fig:gens} provides a visual illustration of the generations within the EP \ref{EPA} algorithm. Each box in figure \ref{fig:gens} illustrates an individual and each row illustrates a population. The green boxes indicate the fittest individual of each population (i.e., lowest AE).

\subsection{Evolutionary programming algorithm}\label{EPA}
The version of evolutionary programming used in this paper follows the algorithm described in \cite{intro:algo}.
\begin{enumerate}
    \item Generate an initial population of individuals. The number of individuals differs between the experiments.
    \item\label{EPA:eval} Evaluate fitness as described in \ref{Defs}.
    \item Select the fittest individual of the population by selecting the individual with lowest AE.
    \item If the fittest individual has an average error better $0.01$, return that individual and exit. $0.01$ was chosen as a way to try to get the AE lower than 1 percent.
    \item Else generate a new population by applying a mutation to the fittest individual. The mutation differs between the experiments but the general principle is to add a random number from a Gaussian distribution to each of the new offspring. The mutation used for the experiments is described in detail in section \ref{experiments}.
    \item Go to step \ref{EPA:eval}.
\end{enumerate}

\begin{algorithm}
\caption{Function used in experiment 1.}
\label{alg:mutate1}
\begin{algorithmic}[!ht]
\Procedure{Mutate}{$value$}
\State $add \sim \mathcal{N}(0,\,0.05)\,$
\While {$value+add<0$}
    \State $add \gets add*0.5$
\EndWhile
\Return $value+add$
\EndProcedure
\end{algorithmic}
\end{algorithm}
\vspace{5mm}

\begin{algorithm}
\caption{Function used in experiment 2 and 3.}
\label{alg:mutate2}
\begin{algorithmic}[1]
\Procedure{Mutate}{$value$}
\State $add \sim value*\mathcal{N}(0,\,0.5)\,$
\While {$value+add<0$}
    \State $add \gets add*0.5$
\EndWhile
\Return $value+add$
\EndProcedure
\end{algorithmic}
\end{algorithm}
\vspace{5mm}

\subsection{Experiments}\label{experiments}
Table \ref{table:ME1} describes all of the experimental configurations. All experiments ran for 100 generations and no experiment reached the set criteria of $AE < 0.01$. All experiments began with the same initial population, which looked like the following:
$k_{pv} \sim \mathcal{U}(0, 1)\,$, 
$k_{iv} \sim \mathcal{U}(0, 0.1)\,$, 
$k_{dv} \sim \mathcal{U}(0, 0.01)\,$,
$k_{pa} \sim \mathcal{U}(0, 1)\,$, 
$k_{ia} \sim \mathcal{U}(0, 0.1)\,$, 
$k_{da} \sim \mathcal{U}(0, 0.01)\,$.

The mutate algorithm provided in \cite{intro:mutate} and used in experiment 1 did not perform well for this application because the parameters differed in size by more than an order of magnitude. Therefore, experiments 2 and 3 used a mutate algorithm that scales the mutation with the value being mutated.

\begin{table}[ht!]
\centering
\begin{tabular}{|c c c|} 
 \hline
 Experiment & Mutation algorithm & Population size \\
 \hline\hline
 1 & See algorithm \ref{alg:mutate1} & 10 \\
 2 & See algorithm \ref{alg:mutate2} & 10 \\
 3 & See algorithm \ref{alg:mutate2} & 20 \\
 \hline
\end{tabular}
\caption{Configurations for the experiments.}
\label{table:ME1}
\end{table}

\section{Result}
For each experiment, this paper will present the AE value as well as a graph illustrating the step response of the PID for both train and test. Table \ref{table:res} lists every experiment's best parameters and lowest AE for train and test.

\begin{table}[H]
\centering
\begin{tabular}{|c c c c c c c|} 
 \hline
 Experiment & Type & $k_p$ & $k_i$ & $k_d$ & AE train & AE Test \\
 \hline\hline
 
 1 & Linear & $7.78*10^{-2}$ & $1.63*10^{-1}$ & 0 & 0.0563 & 0.0707 \\ 
 1 & Angular &  $7.83*10^{-2}$ & $4.24*10^{-2}$ & 0  & 0.0615 & 0.0529 \\
 \hline
 
 2 & Linear & $8.16*10^{-2}$ & $2.12 * 10^{-6}$ & 0 & 0.0312 & 0.0406 \\ 
 2 & Angular & $1.17*10^{-1}$ & $6.34*10^{-6}$ & $2.60*10^{-9}$ & 0.0253 &  0.0414 \\
 \hline
 
 3 & Linear & $1.23*10^{-1}$ & $2.12*10^{-5}$ & 0 & 0.0313 & 0.0514 \\ 
 3 & Angular & $1.21*10^{-1}$ & $3.25*10^{-4}$ & $3.35*10^{-8}$ & 0.0234 & 0.0421 \\
 \hline
 
\end{tabular}
\caption{Results for the experiments.}
\label{table:res}
\end{table}

\begin{figure}[H]
   \centering
       \includegraphics[width=0.70\textwidth]{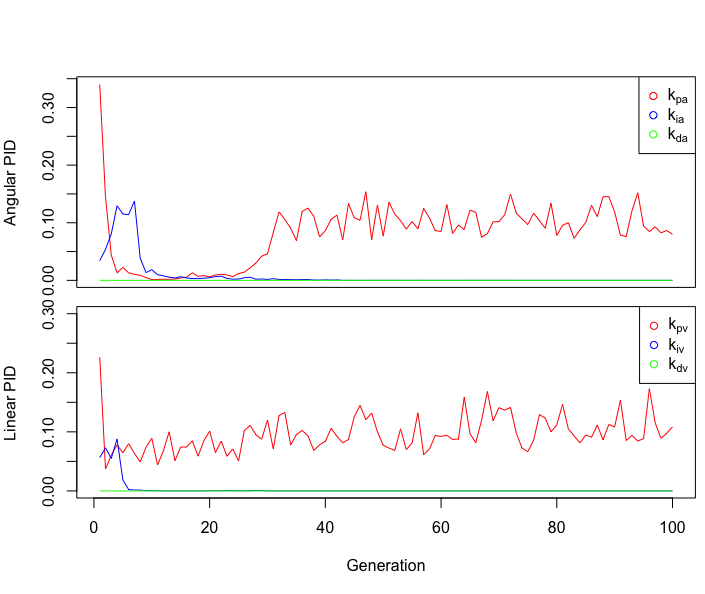}
       \caption{Figure showing PID parameters evolving over generations.}
   \label{fig:RE2:params}
\end{figure}

\begin{figure}[H]
   \centering
       \includegraphics[width=0.70\textwidth]{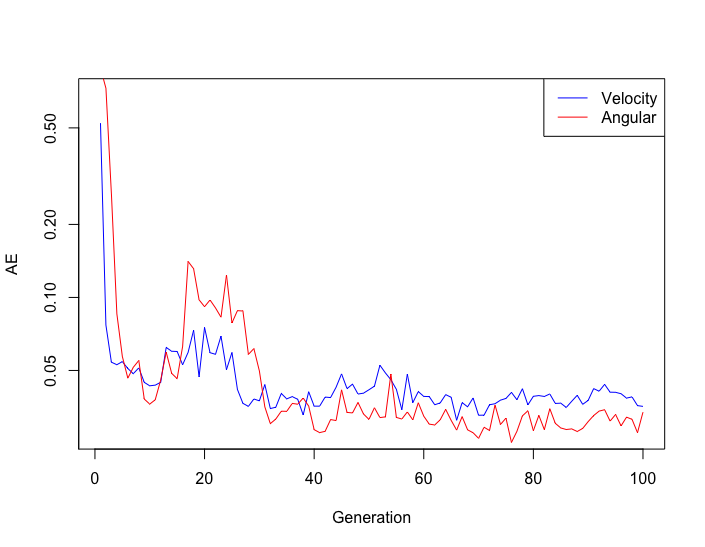}
       \caption{The AE over generations during experiment 2.}
   \label{fig:RE2:gens}
\end{figure}

\begin{figure}[H]
   \centering
       \includegraphics[width=0.70\textwidth]{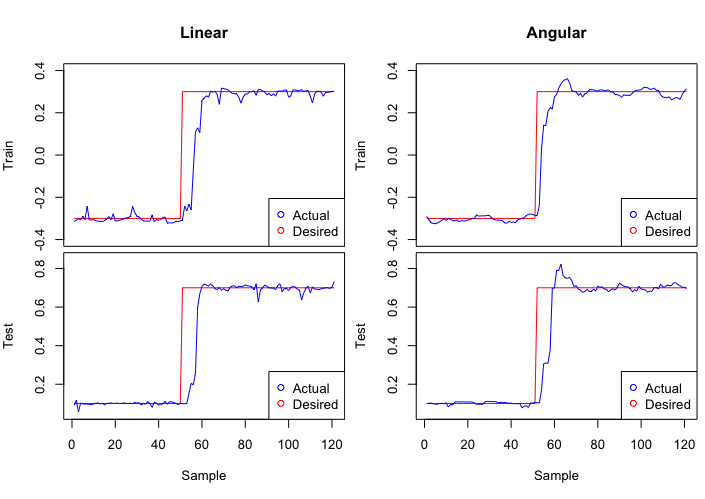}
       \caption{Step response for train and test for both linear and angular velocity.}
   \label{fig:RE2:step}
\end{figure}

In addition to the results presented in Table \ref{table:res}, Figure \ref{fig:RE2:params} illustrates how the PID parameters evolve over every generation in experiment 2. For experiment 2, Figure \ref{fig:RE2:gens} shows the AE across all generations and Figure \ref{fig:RE2:step} shows the step response for train and test for both linear and angular velocity.


\section{Discussion}
Overall, we can conclude that using EP is successful in finding sufficient parameters given enough training time as described in \cite{intro:mutate}. The results presented in this paper are somewhat surprising. As we can see, the d part of PID becomes either 0 or negligible after running the algorithm for a few generations. The same thing happens to the i part, but to an lesser extent. An explanation for this behavior could be the way that the simulator is built. This begs the question of whether the results are useful in an environment outside of our simulation. Finally, we noticed that increasing the population size from 10 to 20 did not significantly improve the results.

\vspace{5mm}

When comparing the results with the results from R.Johns \cite{rasmus} and S.Hadenius \cite{simon}, we can see that they got similar results, including very small ID-parameters. This adds confidence that the algorithm works but that the simulation is not optimal for tuning the PIDs.

\vspace{5mm}

There are several variations to the experiment that could be done to improve the results. One way is to divide the route into different parts and then calculate the AE for each part. That would minimize the risk of a good integral value being punished by the algorithm for a slow rise time. Second, other fitness functions could be used instead of AE, including mean squared error. Third, training the parameters on a dynamic route could take the algorithm out of a local minima. Lastly, running the robot with the best known parameters in between every run would allow each individual to start with as low of an error as possible. In this paper, bad parameters from a previous individual could affect the AE of the next tested individual. 

\vskip 0.2in
\medskip
\printbibliography

\end{document}